\documentclass{article}

\usepackage[preprint,nonatbib]{nips_2018}

\usepackage{url}            % simple URL typesetting
\usepackage{amsfonts}       % blackboard math symbols
\usepackage[square,sort,comma,numbers]{natbib}

\usepackage{color}
\usepackage{enumitem}
\usepackage{graphicx}
\usepackage{multirow}
\usepackage{amsmath}
\usepackage{amssymb}
\usepackage{algorithm}
\usepackage{algpseudocode}

\title{Shape Robust Text Detection with Progressive Scale Expansion Network}

\author{
   Xiang~Li$^1$\thanks{Equal contribution. Please contact xiang.li.implus@njust.edu.cn and wangwenhai362@163.com.}, ~Wenhai Wang$^2$$^1$$^*$, Wenbo Hou$^2$, Ruo-Ze Liu$^2$, Tong Lu$^2$, ~Jian Yang$^1$\\
   $^1$DeepInsight@PCALab, Nanjing University of Science and Technology\\
   $^2$National Key Lab for Novel Software Technology, Nanjing University\\
}

\begin{document}

\maketitle

\begin{abstract}
	The challenges of shape robust text detection lie in two aspects: 1) most existing quadrangular bounding box based detectors are difficult to locate texts with arbitrary shapes, which are hard to be enclosed perfectly in a rectangle; 2) most pixel-wise segmentation-based detectors may not separate the text instances that are very close to each other. To address these problems, we propose a novel Progressive Scale Expansion Network (PSENet), designed as a segmentation-based detector with multiple predictions for each text instance. These predictions correspond to different ``kernels'' produced by shrinking the original text instance into various scales. Consequently, the final detection can be conducted through our progressive scale expansion algorithm which gradually expands the kernels with minimal scales to the text instances with maximal and complete shapes. Due to the fact that there are large geometrical margins among these minimal kernels, our method is effective to distinguish the adjacent text instances and is robust to arbitrary shapes. The state-of-the-art results on ICDAR 2015 and ICDAR 2017 MLT benchmarks further confirm the great effectiveness of PSENet. Notably, PSENet outperforms the previous best record by absolute 6.37\% on the curve text dataset SCUT-CTW1500. Code will be available in https://github.com/whai362/PSENet.
\end{abstract}

\section{Introduction}
Recently, natural scene text detection has attracted extensive attention for its numerous applications, such as scene understanding, product identification, automatic driving and target geolocation. However, due to the large variations in foreground texts and background objects, and the diverse text variabilities in shapes, colors, fonts, orientations and scales, along with the extreme illumination and occlusion, text detection in natural scene is still faced with considerable challenges.

\begin{figure}
	\centering
	\setlength{\fboxrule}{0pt}
	\fbox{\includegraphics[width=1\textwidth]{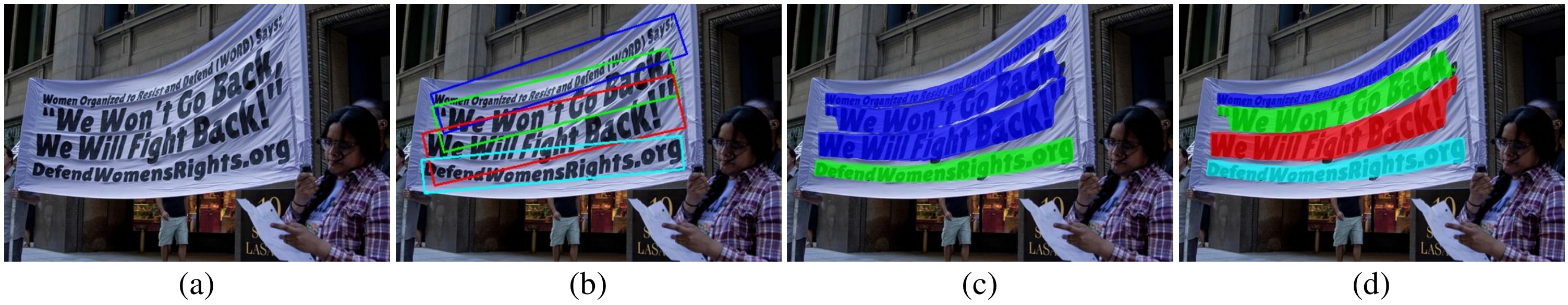}}
	\vspace{-14pt}
	\caption{The results of different methods, best viewed in color. (a) is the original image. (b) refers to the result of bounding box regression-based method, which displays disappointing detections as the red box covers nearly more than half of the context in the green box. (c) is the result of semantic segmentation, which mistakes the 3 text instances for 1 instance since their boundary pixels are partially connected. (d) is the result of our proposed PSENet, which successfully distinguishs and detects the 4 unique text instances.}
	\label{fig:diff-meth-res}
  \vspace{-8pt}
\end{figure}

Nevertheless, great progress has been made in recent years with the amazing development of Convolutional Neural Networks (CNNs) \cite{he2016deep,huang2017densely,ren2015faster}. Based on bounding box regression, a list of methodologies \cite{tian2016detecting, zhou2017east, shi2017detecting, jiang2017r2cnn, zhong2016deeptext, liu2018fots, he2017single, hu2017wordsup, lyu2018multi} has been proposed to successfully locate the text targets in forms of rectangles or quadrangles with certain orientations. Unfortunately, these frameworks cannot detect the text instances with arbitrary shapes (e.g., the curve texts), which also often appear in natural scenes (see Fig. \ref{fig:diff-meth-res} (b)). Naturally,
semantic segmentation-based methods can be taken into consideration to explicitly handle the curve text detection problems. Although pixel-wise segmentation can extract the regions of arbitrary-shaped text instances, it may still fail to separate two text instances when they are relatively close, because their shared adjacent boundaries will probably merge them together as one single text instance (see Fig. \ref{fig:diff-meth-res} (c)).

To address these problems, in this paper, we propose a novel instance segmentation network, namely, Progressive Scale Expansion Network (PSENet). There are two advantages of the proposed PSENet. Firstly, as a segmentation-based method, PSENet is able to locate texts with arbitrary shapes. Secondly, we put forward a progressive scale expansion algorithm, with which the closely adjacent text instances can be identified successfully (see Fig. \ref{fig:diff-meth-res} (d)). Specifically, we assign each text instance with multiple predicted segmentation areas. For convenience, we denote these segmentation areas as ``kernels'' in this paper and for one text instance, there are several corresponding kernels. Each of the kernels shares the similar shape with the original entire text instance, and they all locate at the same central point but differ in scales. To obtain the final detections, we adopt the progressive scale expansion algorithm. It is based on Breadth-First-Search (BFS) and is composed of 3 steps: 1) starting from the kernels with minimal scales (instances can be distinguished in this step); 2) expanding their areas by involving more pixels in larger kernels gradually; 3) finishing until the largest kernels are explored.

The motivations of the progressive scale expansion are mainly of four folds. Firstly, the kernels with minimal scales are quite easy to be separated as their boundaries are far away from each other. Therefore, it overcomes the major drawbacks of the previous segmentation-based methods; Secondly, the largest kernels or the complete areas of text instances are indispensable for achieving the final precise detections; Thirdly, the kernels are gradually growing from small to large scales, and thus the smoonth surpervisions would make the networks much easier to learn; Finally, the progressive scale expansion algorithm ensures the accurate locations of text instances as their boundaries are expanded in a careful and gradual manner.

To show the effectiveness of our proposed PSENet, we conduct extensive experiments on three competitive benchmark datasets including ICDAR 2015 \cite{karatzas2015icdar}, ICDAR 2017 MLT \cite{icdar2017mlt} and SCUT-CTW1500 \cite{Liu2017Detecting}. 
Among these datasets, SCUT-CTW1500 is explicitly designed for curve text detection, and on this dataset we surpass the previous state-of-the-art result by absolute 6.37\%.
Furthermore, the proposed PSENet achieves better or at least comparable performance on the ordinary quadrangular text datasets: ICDAR 2015 and ICDAR 2017 MLT, when compared with the existing state-of-the-art methods. 

%{\color{red}{Comparing with the existing state-of-the-art methods, the proposed PSENet achieve better or at least comparable performance on ICDAR 2015 [*] and ICDAR 2017 MLT [*], and further pushes the Hmean on SCUT-CTW1500 [*] and Total-Text [*] to * and *.}}

The main contributions of this paper are as follows:
\begin{itemize}[leftmargin=*]
\item We propose a novel Progressive Scale Expansion Network (PSENet) which can precisely detect text instances with arbitrary shapes.
\item We propose a progressive scale expansion algorithm which is able to accurately separate the text instances standing closely to each other.
\item Our proposed PSENet significantly surpasses the state-of-the-art methods on the curve text detection dataset SCUT-CTW1500. Furthermore, it also achieves competitive results on the regular quadrangular text benchmarks: ICDAR 2015 and ICDAR 2017 MLT. 
\end{itemize}

\section{Related Work}
Text detection has been an active research topics in computer vision for a long period of time. \cite{zhong2016deeptext, liao2017textboxes} successfully adopted the pipelines of object detection into text detection and obtained good performance on horizontal text detection.
After that, \cite{shi2017detecting, zhou2017east, liu2018fots, jiang2017r2cnn, he2017single, hu2017wordsup} took the orientation of text line into consideration and made it possible to detect arbitrary-oriented text instances. Recently, \cite{lyu2018multi} utilized corner localization to find suitable irregular quadrangles for text instances. The detection manners are evolving from horizontal rectangle to rotated rectangle and further to irregular quadrangle. However, besides the quadrangular shape, there are many other shapes of text instances in natural scene. Therefore, some researches began to explore curve text detection and obtained certain results. \cite{Liu2017Detecting} tried to regress the relative positions for the points of a 14-sided polygon. \cite{zhu2018sliding} detected curve text by locating two end points in the sliding line which slides both horizontally and vertically. A fused detector was proposed in \cite{dai2017fused} based on bounding box regression and semantic segmentation. However, since their current performances are not very satisfied, there is still a large space for promotion in curve text detection, and the detectors for arbitrary-shaped texts still need more explorations.

\section{Proposed Method}
In this section, we first introduce the overall pipeline of the proposed Progressive Scale Expansion Network (PSENet). Next, we present the details of progressive scale expansion algorithm, and show how it can effectively distinguish the adjacent text instances. Further, the way of generating label and the design of loss function are introduced. At last, we describe the implementation details of PSENet.

\subsection{Overall Pipeline}
\begin{figure}
  \centering
  \setlength{\fboxrule}{0pt}
  \fbox{\includegraphics[width=1\textwidth]{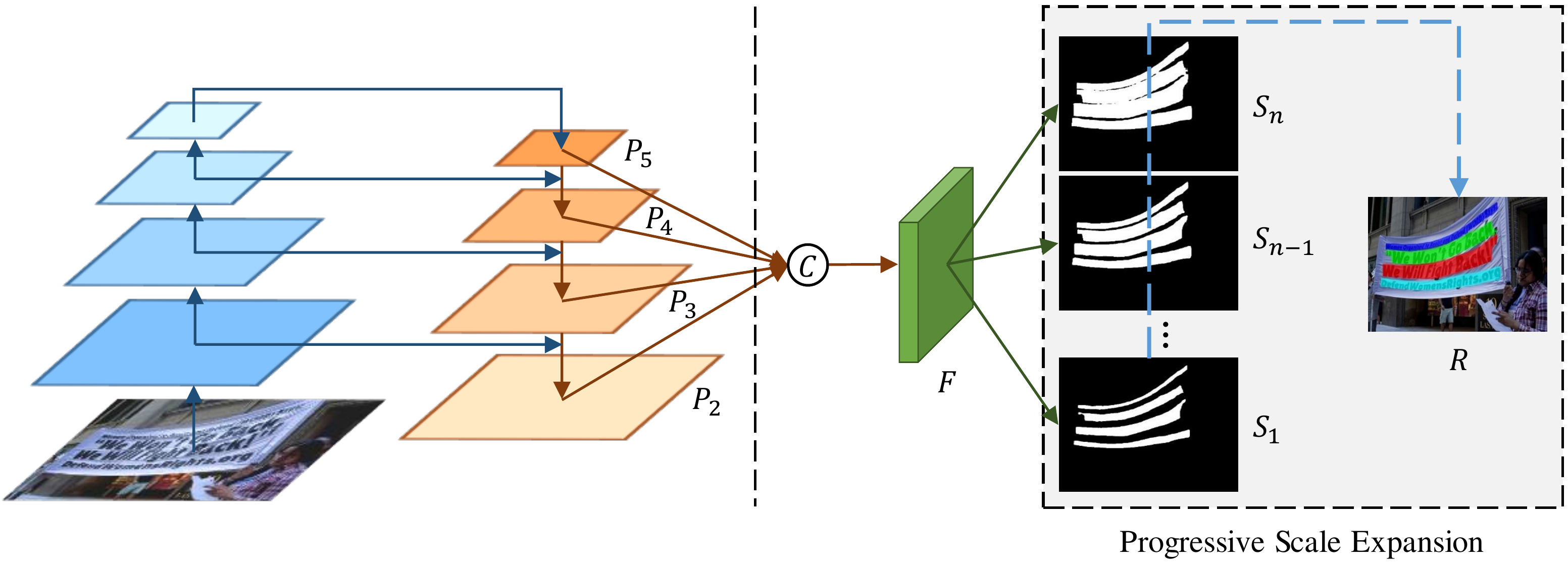}}
  \vspace{-14pt}
  \caption{Illustration of our overall pipeline. The left part is implemented from FPN \cite{lin2017feature}. The right part denotes the feature fusion and the progressive scale expansion algorithm.}
  \label{fig:pipeline}
  \vspace{-6pt}
\end{figure}

The overall pipeline of the proposed PSENet is illustrated in Fig. \ref{fig:pipeline}. 
% We use ResNet \cite{he2016identity} as the backbone of PSENet. 
Inspired by FPN \cite{lin2017feature}, we concatenate low-level feature maps with high-level feature maps and thus have four concatenated feature maps. These maps are further fused in $F$ to encode informations with various receptive views. Intuitively, such fusion is very likely to facilitate the generations of the kernels with various scales. Then the feature map $F$ is projected into $n$ branches to produce multiple segmentation results $S_1, S_2, ..., S_n$. Each $S_i$ would be one segmentation mask for all the text instances at a certain scale. The scales of different segmentation mask are decided by the hyper-parameters which will be discussed in Sec. \ref{sec:label-gen}.
% The scales vary in a range which is decided by the hyper-parameters {\color{blue} mentioned in Sec. *}. 
Among these masks, $S_1$ gives the segmentation result for the text instances with smallest scales (i.e., the minimal kernels) and $S_n$ denotes for the original segmentation mask (i.e., the maximal kernels). After obtaining these segmentation masks, we use progressive scale expansion algorithm to gradually expand all the instances' kernels in $S_1$, to their complete shapes in $S_n$, and obtain the final detection results as $R$.

%{\color{red}{Using feature map $F$, we make binary classification pixel-wisely and obtain $n$ segmentation results $S_0, S_1, ..., S_n$} for $n$ different scale text lines.} $S_0$ is the segmentation result for {\color{red}{the smallest scale text lines}}. It is obvious that the text lines in $S_0$ can be well separate and thus each connected component in $S_0$ can be viewed as a kernel for a text line. {\color{red}{With the increasing of $i$, the scale of text lines in $S_i$ is expanded progressively and $S_n$ is the segmentation result for complete text lines.}} At last, we use progressive scale expansion algorithm to gradually expand the kernels in $S_0$ to a complete text lines in $S_n$ and get the final text lines in $R$. The detail of PSENet is shown in Tab. \ref{tab: pipeline}. 
%{\color{red}{Do we need to discribe the table or the overflow of PSENet?}}

\iffalse
\begin{table}
	\scriptsize
	% \small
	\centering
	\renewcommand\arraystretch{1.2}
	\newcommand{\tabincell}[2]{\begin{tabular}{@{}#1@{}}#2\end{tabular}}
	\caption{PSENet.}
	% \vspace{-10pt}
	\scalebox{1}{
		\begin{tabular}{c|c|c|c}
			\hline
			Module & Operation & Output Channels & Output Name \\
			\hline
			FPN Backbone & Implement from FPN & $256$ & $P_2, P_3, P_4, P_5$ \\
			\hline
			$C(\cdot)$ & $P_2 \parallel Up_{\times 2}(P_3) \parallel Up_{\times 4}(P_4) \parallel Up_{\times 8}(P_5)$ & $1024$ & $F$ \\
			\hline
			Conv-BN-ReLU & $3 \times 3$ conv, stride $1$, padding $1$; BN; ReLU & $256$  & - \\
			\hline
			Conv-Up-Sigmoid & $1 \times 1$ conv, stride $1$, padding $0$; Upsample($\times 4$); Sigmoid & $1$ & $S_1, S_2, ..., S_n$ \\
			\hline
			PSE & Progressive Scale Expansion & $1$ & $R$ \\
			\hline
		\end{tabular}}
	\label{tab: pipeline}
	% \vspace{-10pt}
\end{table}
\fi

\subsection{Progressive Scale Expansion Algorithm}
As shown in Fig. \ref{fig:diff-meth-res} (c), it is hard for segmentation-based method to separate the text instances that are close to each other. To solve this problem, we propose the progressive scale expansion algorithm. 

%It involves two factors in general: 1) the first is the multi-branch output architecture at the end of the body network; 2) the second is the BFS-based algorithm that utilizes the multiple segmentation predictions to decide the final detections. Since the multi-branch design is clearly demonstrated in Fig. \ref{fig:pipeline}, we focus on the second one -- the BFS-based algorithm in the following description. For clarity, we name this algorithm as progressive scale expansion algorithm.

\begin{figure}
  \centering
  \setlength{\fboxrule}{0pt}
  \fbox{\includegraphics[width=1\textwidth]{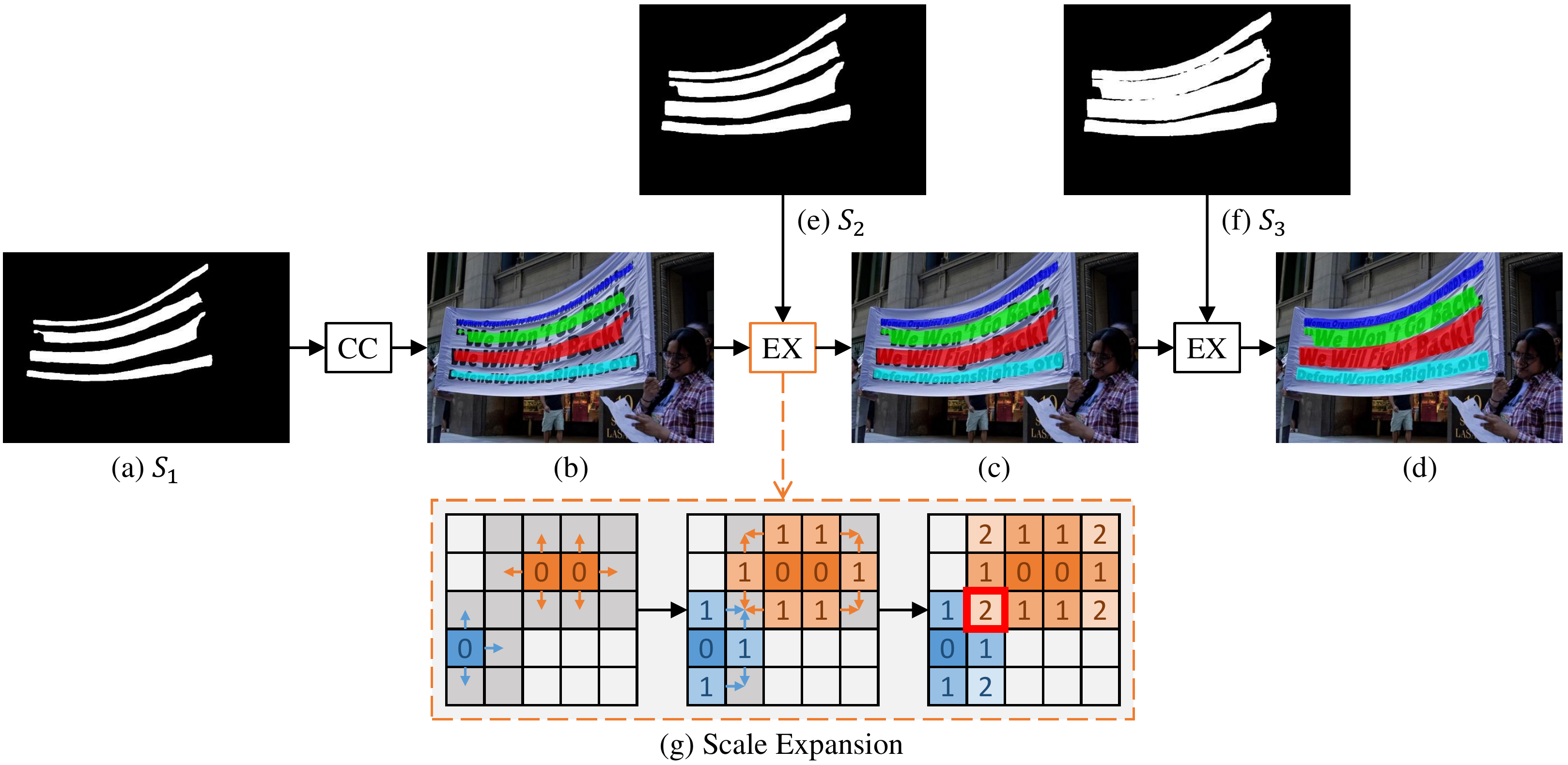}}
  \vspace{-14pt}
  \caption{The procedure of progressive scale expansion algorithm. CC refers to the function of finding connected components. EX represents the scale expansion algorithm. (a), (e) and (f) refer to $S_1$, $S_2$ and $S_3$, respectively. (b) is the initial connected components. (c) and (d) is the results of expansion. (g) shows the illustration of expansion. The red box in (g) refers to the conflicted pixel.}
  \label{fig:pse}
    \vspace{-4pt}
\end{figure}

Here is a vivid example (see Fig. \ref{fig:pse}) to explain the procedure of progressive scale expansion algorithm, whose central idea is brought from the Breadth-First-Search (BFS) algorithm. In the example, we have $3$ segmentation results $S = \{S_1, S_2, S_3\}$ (see Fig. \ref{fig:pse} (a), (e), (f)). At first, based on the minimal kernels' map $S_1$ (see Fig. \ref{fig:pse} (a)), 4 distinct connected components $C = \{c_1, c_2, c_3, c_4\}$ can be found as initializations. The regions with different colors in Fig. \ref{fig:pse} (b) represent these different connected components, respectively. By now we have all the text instances' central parts (i.e., the minimal kernels) detected. 
Then, we progressively expand the detected kernels by merging the pixels in $S_2$, and then in $S_3$. The results of the two scale expansions are shown in Fig. \ref{fig:pse} (c) and Fig. \ref{fig:pse} (d), respectively.
Finally, we extract the connected components which are marked with different colors in Fig. \ref{fig:pse} (d) as the final predictions for text instances.

The procedure of scale expansion is illustrated in Fig. \ref{fig:pse} (g). The expansion is based on Breadth-First-Search algorithm which starts from the pixels of multiple kernels and iteratively merges the adjacent text pixels. Note that there may be conflicted pixels during expansion, as shown in the red box in Fig. \ref{fig:pse} (g). The principle to deal with the conflict in our practice is that the confusing pixel can only be merged by one single kernel on a first-come-first-served basis. Thanks to the ``progressive'' expansion procedure, these boundary conflicts will not affect the final detections and the performances.
The detail of scale expansion algorithm is summarized in Algorithm \ref{alg:expansion}. In the pseudocode, $T, P$ are the intermediate results. $Q$ is a queue. $\mbox{Neighbor}(\cdot)$ represents the neighbor pixels of $p$. $\mbox{GroupByLabel}(\cdot)$ is the function of grouping the intermediate result by label. ``$S_i[q] = \mbox{True}$'' means that the predicted value of pixel $q$ in $S_i$ belongs to the text part.

%which starts from the kernels of text lines and expand the kernels to complete text lines by degrees. 
%Fig. \ref{fig:pse} illustrates the example of progressive scale expansion algorithm. In the example, we let $n = 3$ and get $3$ segmentation results $S_0$, $S_1$, $S_2$. 
%Firstly, we find the connected components $C$ in $S_0$ (Fig. \ref{fig:pse}(a)). The regions with different colors in Fig. \ref{fig:pse}(b) represent the different connected components and each connected component $c_i \in C$ can be viewed as a kernel for a text line. 
%Next, we iteratively expand each kernel $c_i \in C$ according to the follow-on segmentation results $R_1$ (Fig. \ref{fig:pse}(e)) and $R_2$ (Fig. \ref{fig:pse}(f)). The expansion results are shown in Fig. \ref{fig:pse}(c) and Fig. \ref{fig:pse}(d), respectively.
%{\color{red}{Finally, we extract the regions with different colors in Fig. \ref{fig:pse}(d) as the final text lines.}}

%Fig. \ref{fig:pse}(g) shows the procedure of expansion. The expansion is based on breadth-first search algorithm which starts from the pixels of different kernels and iteratively merges the adjacent text pixels. The procedure is summarized in Algorithm \ref{alg: expansion}.
% The process of expansion starts with the kernels from previous step and me
% The process of expansion is shown in the blue dashed box in Fig. \ref{fig:pse}.

% \iffalse
\begin{algorithm}[t]
	\scriptsize
    \caption{Scale Expansion Algorithm}
    \begin{algorithmic}[1]
        \Require Kernels: $C$, Segmentation Result: $S_i$
        \Ensure Scale Expanded Kernels: $E$
        \Function {Expansion}{$C$, $S_i$}
            \State $T \gets \emptyset; P \gets \emptyset; Q \gets \emptyset$
            %\State $$
            % \State $Q \gets \emptyset$
            \For{each $c_i \in C$}
           		\State $T \gets T \cup \{(p, label) \mid (p, label) \in c_i\}; P \gets P \cup \{p \mid (p, label) \in c_i\}$
           		\State $\mbox{\bf{Enqueue}}(Q, c_i)$ \ \ \ \ \ \ \ \ \ \ \ \ \ \ \ \ \ \ \ \ \ \ \ \ \ \ \ \ \ \ \ // push all the elements in $c_i$ into $Q$
           		%\State $ Q \gets Q \cup \{(p, label) \mid (p, label) \in c_i\}$ 
           		%\State $$
            \EndFor
            \While{$Q \neq \emptyset$}
                \State $(p, label) \gets \mbox{\bf{Dequeue}}(Q)$   \ \ \ \ \ \ \ \ \ \ \ \ // pop the first element of $Q$
                \If{$\exists q \in \mbox{\bf{Neighbor}}(p)$ and $q \notin P$ and $S_i[q] = \mbox{True}$}
                	\State $T \gets T \cup \{(q, label)\}; P \gets P \cup \{q\}$
                	\State $\mbox{\bf{Enqueue}}(Q, (q, label))$ \ \ \ \ \ \ \ \ \ // push the element $(q, label)$ into $Q$
                \EndIf
                % \State $Q \gets Q - \{(p, label)\}$
            \EndWhile
            \State $E = \mbox{\bf{GroupByLabel}}(T)$
            \State \Return{$E$}
        \EndFunction
    \end{algorithmic}
    \label{alg:expansion}
\end{algorithm}
% \fi
\subsection{Label Generation}
\label{sec:label-gen}
\begin{figure}[t]
  \centering
  \setlength{\fboxrule}{0pt}
  \fbox{\includegraphics[width=0.8\textwidth]{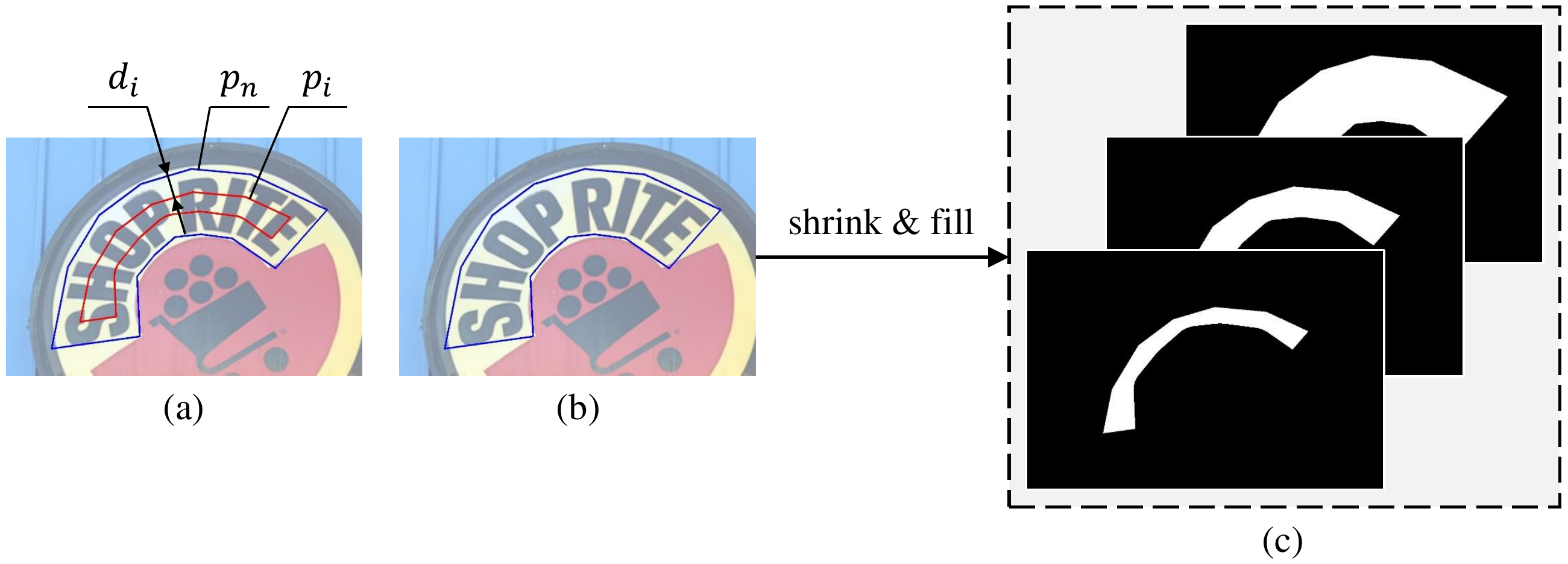}}
  \vspace{-8pt}
  \caption{The illustration of label generation. (a) contains the annotations for $d$, $p_i$ and $p_n$. (b) shows the original text instances. (c) shows the segmentation masks with different kernel scales.}
  \label{fig:label_gen}

\end{figure}

As illustrated in Fig. \ref{fig:pipeline}, PSENet produces segmentation results (e.g. $S_1, S_2, ..., S_n$) with different kernel scales. Therefore, it requires the corresponding ground truths with different kernel scales as well during training. In our practice, these ground truth labels can be conducted simply and effectively by shrinking the original text instance.
% shrinking the original segmentation areas for all the text instances. 
The polygon with blue border in Fig. \ref{fig:label_gen} (b) denotes the original text instance % area for the text instance 
and it corresponds to the largest segmentation label mask (see the rightmost map in Fig. \ref{fig:label_gen} (c)). To obtain the shrunk masks sequentially in Fig. \ref{fig:label_gen} (c), we utilize the Vatti clipping algorithm \cite{vatti1992generic} to shrink the original polygon $p_n$ by $d_i$ pixels and get shrunk polygon $p_i$ (see Fig. \ref{fig:label_gen} (a)). Subsequently, each shrunk polygon $p_i$ is transferred into a 0/1 binary mask for segmentation label ground truth. We denote these ground truth maps as $G_1, G_2, ..., G_n$ respectively. Mathematically, if we consider the scale ratio as $r_i$, the margin $d_i$ between $p_n$ and $p_i$ can be calculated as:
%The blue polygon in Fig. \ref{fig:label_gen}(b) is the original text line and the white filled polygons in Fig. \ref{fig:label_gen}(c) are shrunk text line ground truths ($e.g.,  G_0, G_1, ..., G_n$) with different scales. Let us consider the shrink rate $r$, the margin $d$ between original polygon $p$ (blue polygon in Fig. \ref{fig:label_gen}(a)) and shrunk polygon $p_s$ (red polygon in Fig. \ref{fig:label_gen}(a)) can formulate as follows:
\begin{equation}
	d_i = \frac{\mbox{Area}(p_n) \times (1 - r_i^2)}{\mbox{Perimeter}(p_n)},
	\label{eqn:d_i}
\end{equation}
where $\mbox{Area}(\cdot)$ is the function of computing the polygon area, $\mbox{Perimeter}(\cdot)$ is the function of computing the polygon perimeter. Further, we define the scale ratio $r_i$ for ground truth map $G_i$ as:
\begin{equation}
	r_i = 1 - \frac{(1 - m) \times (n - i)}{n - 1},
	\label{eqn:r_i}
\end{equation}
where $m$ is the minimal scale ratio, which is a value in $(0, 1]$. Based on the definition in Eqn.~\eqref{eqn:r_i}, the values of scale ratios (i.e., $r_1, r_2, ..., r_n$) are decided by two hyper-parameters $n$ and $m$, and they increase linearly from $m$ to $1$.

%During generating label, we first use Vatti clipping algorithm [*] shrink the original polygon $p$ inside $d$ pixels and get shrunk polygon $p_s$. Then, we fill the shrunk polygon $p_s$ with label $1$ and the background with label $0$. {\color{red}{Note that, we use package...}}

\subsection{Loss Function}
For learning PSENet, the loss function can be formulated as:
\begin{equation}
	L =  \lambda L_c + (1 - \lambda) L_s,
	\label{eqn:loss-tot}
\end{equation}
where $L_c$ and $L_s$ represent the losses for the complete text instances and the shrunk ones respectively, and $\lambda$ balances the importance between $L_c$ and $L_s$. 

It is common that the text instances usually occupy only an extremely small region in natural images, which makes the predictions of network bias to the non-text region, when binary cross entropy \cite{de2005tutorial} is used. Inspired by \cite{milletari2016v}, we adopt dice coefficient in our experiment. The dice coefficient $D(S_i, G_i)$ is formulated as in Eqn.~\eqref{eqn:dice-coef}:
\begin{equation}
	D(S_i, G_i) = \frac{2 \sum\nolimits_{x,y} (S_{i, x, y} * G_{i, x, y})}{\sum\nolimits_{x, y} S_{i, x, y}^2 + \sum\nolimits_{x, y} G_{i, x, y}^2},
	\label{eqn:dice-coef}
\end{equation}
where $S_{i, x, y}$ and $G_{i, x, y}$ refer to the value of pixel $(x, y)$ in segmentation result $S_i$ and ground truth $G_i$, respectively. 

Furthermore, there are many patterns similar to text strokes, such as fences, lattices, etc. Therefore, we adopt Online Hard Example Mining (OHEM) \cite{shrivastava2016training} to $L_c$ during training to better distinguish these patterns.

$L_c$ focuses on segmenting the text and non-text region. Let us consider the training mask given by OHEM as $M$, and thus $L_c$ can be written as:
\begin{equation}
	L_c = 1 - D(S_n \cdot M, G_n \cdot M),
	\label{eqn:loss-complete}
\end{equation}
$L_s$ is the loss for shrunk text instances. Since they are encircled by the original areas of the complete text instances, we ignore the pixels of non-text region in the segmentation result $S_n$ to avoid a certain redundancy. Therefore, $L_s$ can be formulated as follows:
\begin{equation}
	L_s = 1 - \frac{\sum\nolimits_{i = 1}^{n - 1} D(S_i \cdot W, G_i \cdot W)}{n - 1},\ \ \ \  W_{x, y} = 
	\left\{
	\begin{array}{ll}
	1, & if \ S_{n, x, y} \ge 0.5; \\  
	0, & otherwise.   
	\end{array}
	\right.  
	\label{eqn:loss-shrink}
\end{equation}

Here, $W$ is a mask which ignores the pixels of non-text region in $S_n$, and $S_{n, x, y}$ refers to the value of pixel $(x, y)$ in $S_n$.

\subsection{Implementation Details}
% {\color{red}{to do}}
The backbone of PSENet is implemented from FPN \cite{lin2017feature}. We firstly get four $256$ channels feature maps (i.e. $P_2, P_3, P_4, P_5$) from the backbone. To further combine the semantic features from low to high levels, we fuse the four feature maps to get feature map $F$ with $1024$ channels via the function $C(\cdot)$ as: $
F = C(P_2, P_3, P_4, P_5) = P_2 \parallel \mbox{Up}_{\times 2}(P_3) \parallel \mbox{Up}_{\times 4}(P_4) \parallel \mbox{Up}_{\times 8}(P_5),
$
where ``$\parallel$'' refers to the concatenation and $\mbox{Up}_{\times 2}(\cdot)$, $\mbox{Up}_{\times 4}(\cdot)$, $\mbox{Up}_{\times 8}(\cdot)$ refer to 2, 4, 8 times upsampling, respectively. Subsequently, $F$ is fed into Conv($3, 3$)-BN-ReLU layers and is reduced to $256$ channels. Next, it passes through multiple 
Conv($1, 1$)-$\mbox{Up}$-Sigmoid layers and produces $n$ segmentation results $S_1, S_2, ..., S_n$. Here, Conv, BN, ReLU and $\mbox{Up}$ refer to convolution \cite{lecun1998gradient}, batch normalization \cite{ioffe2015batch}, rectified linear units \cite{glorot2011deep} and upsampling.

We set $n$ to $6$ and $m$ to $0.5$ for label generation and get the scales $\{0.5, 0.6, 0.7, 0.8, 0.9, 1.0\}$. During training, we ignore the blurred text regions labeled as “DO NOT CARE” in all datasets. The $\lambda$ of loss balance is set to $0.7$. The negative-positive ratio of OHEM is set to 3. The data augmentation for training data is listed as follows: 1) the images are rescaled with ratio $\{0.5, 1.0, 2.0, 3.0\}$ randomly; 2) the images are horizontally fliped and rotated in range $[-10^\circ, 10^\circ]$ randomly; 3) $640 \times 640$ random samples are cropped from the transformed images; 4) the images are normalized using the channel means and standard deviations. 
For quadrangular text dataset, we calculate the minimal area rectangle to extract the bounding boxes as final predictions. For curve text dataset, the Ramer-Douglas-Peucker algorithm \cite{ramer1972iterative} is applied to generate the bounding boxes with arbitrary shapes. 

\section{Experiment}
In this section, we first conduct ablation studies for PSENet. Then, we evaluate the proposed PSENet on three recent challenging public benchmarks: ICDAR 2015, ICDAR 2017 MLT and SCUT-CTW1500 and compare PSENet with many state-of-the-art methods.

\subsection{Benchmark Datasets}
% \paragraph{ICDAR 2015}
{\bf{ICDAR 2015}} (IC15) \cite{karatzas2015icdar} is a commonly used dataset for text detection. It contains a total of 1500 pictures, 1000 of which are used for training and the remaining are for testing. The text regions are annotated by 4 vertices of the quadrangle.
% The data augmentation scheme we used is listed as follows: (a) images are rescaled with ratio $\{0.5, 1.0, 2.0, 3.0\}$ randomly; (b) images are horizontal fliped and rotated in range $[10^\circ, 10^\circ]$ randomly; (c) $640 \times 640$ random samples are cropped from the transformed images; (d) The images are normalized using the channel means and standard deviations. Note that, we ignore the blurred text regions labeled as “DO NOT CARE” in the dataset.

% \paragraph{ICDAR 2017 MLT}
{\bf{ICDAR 2017 MLT}} (IC17-MLT) \cite{icdar2017mlt} is a large scale multi-lingual text dataset, which includes 7200 training images, 1800 validation images and 9000 testing images. The dataset is composed of complete scene images which come from 9 languages. Similarly with ICDAR 2015, the text regions in ICDAR 2017 MLT are also annotated by 4 vertices of the quadrangle.
% We adopt the same data augmentation scheme with ICDAR 2015 to this dataset and also ignore the blurred text regions labeled as “DO NOT CARE”.

% \paragraph{SCUT-CTW1500}
{\bf{SCUT-CTW1500}} is a challenging dataset for curve text detection, which is constructed by Yuliang et al. \cite{Liu2017Detecting}. It consists of 1000 training images and 500 testing images. Different from traditional text datasets (e.g., ICDAR 2015, ICDAR 2017 MLT), the text instances in SCUT-CTW1500 are labelled by a polygon with 14 points which can discribe the shape of an arbitrarily curve text.
% The data augmentation scheme of SCUT-CTW1500 is also the same with ICDAR 2015.

\subsection{Training}
We use the FPN with ResNet \cite{he2016identity} pre-trained on ImageNet dataset \cite{deng2009imagenet} as our backbone. All the networks are trained by using stochastic gradient descent (SGD).
In the experiments on ICDAR datasets, we use 1000 IC15 training images, 7200 IC17-MLT training images and 1800 IC17-MLT validation images to train the model and report the precision, recall and F-measure on the test set of both datasets at the end of training. We use batch size 16 and train the models for 300 epochs. The initial learning rate is set to $1\times10^{-3}$, and is divided by 10 at 100 and 200 epoch. 
On SCUT-CTW1500, we use the 1000 training images to fine-tune the model from the trained model for ICDAR datasets for 400 epochs. The batch size is set to 16 for fine-tuning. The initial learning rate is set to $10^{-4}$, and is divided by 10 at 200 epoch. At the end of fine-tuning, we report the precision, recall and F-measure on the test set.
We use a weight decay of $5 \times 10^{-4}$ and a Nesterov momentum \cite{sutskever2013importance} of 0.99 without dampening. We adopt the weight initialization introduced by \cite{he2015delving}.

\subsection{Ablation Study}

\begin{figure}[t]
  \centering
  \setlength{\fboxrule}{0pt}
  \fbox{\includegraphics[width=0.8\textwidth]{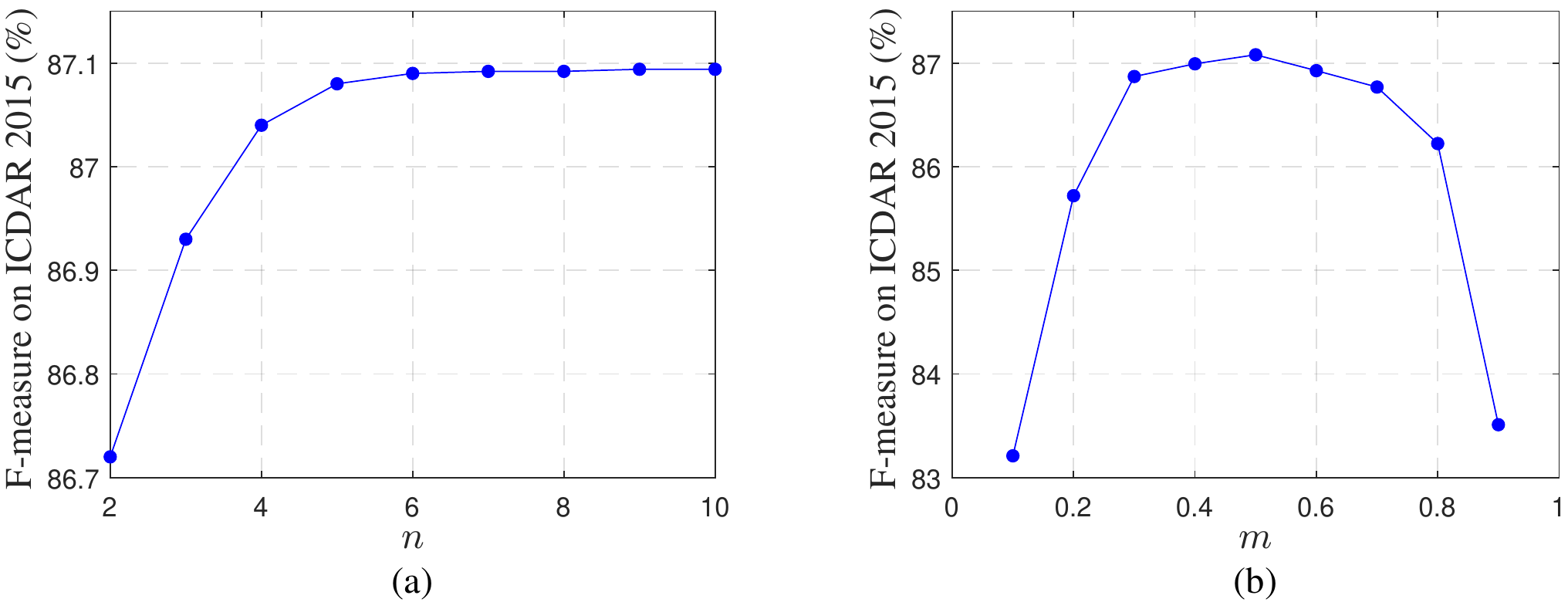}}
  \vspace{-8pt}
  \caption{Ablation study on $m$ and $n$ (Eqn.~\eqref{eqn:r_i}). These results are based on PSENet-1s (Table \ref{tab:ic-ctw}).}
  \label{fig:abs}
  %\vspace{-10pt}
\end{figure}

%\subsubsection{Ablation Study for The Number of Kernel Scales}
\textbf{Why are the multiple kernel scales necessary?} To answer this question, we investigate the effect of the number of scales $n$ on the performance of PSENet. Specifically, we hold the minimal scale $m$ constant and train PSENet with different $n$. In details, we set $m$ to $0.5$ and let $n$ increase from $2$ to $10$. The models are evaluated on ICDAR 2015 dataset. Fig. \ref{fig:abs} (a) shows the experimental results, from which we can find that with the growing of $n$, the F-measure on the test set keeps rising and begins to level off when $n \ge 6$. The informative result suggests that the design of multiple kernel scales is essential and effective, and we also donot need too many scales for the purpose of the efficiency. The original ablation study for $n$ starts from $n = 2$ and fixes $m = 0.5$. Additionally, there is an extreme case when $n = 1$ and $m = 1$, which means we only use the traditional semantic segmentation method to deal with this task. Here we conduct the experiment by setting $n = 1$ and $m = 1$ to serve as a baseline with only one segmentation mask result for predictions. Table \ref{tab:ab} shows the huge performance gap between these two settings, and it further validate the effectiveness of the design of multiple kernel scales.

%\subsubsection{Ablation Study for The Minimal Kernel Scale}
\textbf{How minimal can these kernels be?} We then study the effect of the minimal scale $m$ by setting the number of scales $n$ to $6$ and let the minimal scale $m$ vary from $0.1$ to $0.9$. The models are also evaluated on ICDAR 2015 dataset. We can find from Fig. \ref{fig:abs} (b) that the F-measure on the test set drops when $m$ is too large or too small. When $m$ is too large, it is hard for PSENet to separate the text instances standing closely to each other. When $m$ is too small, PSENet often splits a whole text line into different parts incorrectly and the training can not converge very well.

\begin{table}[t]
	%\scriptsize
	\small
	\centering
	\renewcommand\arraystretch{1.2}
	\newcommand{\tabincell}[2]{\begin{tabular}{@{}#1@{}}#2\end{tabular}}
	\caption{Comparison to the traditional semantic segmentation baseline with the same backbone on ICDAR 2015. ``P'', ``R'', ``F'' refer to Precision, Recall, F-measure respectively.}
	\vspace{-2pt}
	\scalebox{1}{
		\begin{tabular}{l|c|c|c}
			\hline
			{Method}                 &   P (\%)   &  R (\%)    & F (\%)\\
			\hline
			PSENet-1s ($n = 1, m = 1.0,$ semantic segmentation baseline)& 77.41 & 61.53 & {68.56}  \\
			\hline
			PSENet-1s ($n = 6, m = 0.5$)& 88.71 & 85.51 & \textbf{87.08}  \\
			\hline
		\end{tabular}
	}
	\label{tab:ab}
	% \vspace{-10pt}
\end{table}

% \subsubsection{Ablation Study for Network}
% We then 

\subsection{Comparisons with State-of-the-Art Methods}
\textbf{Detecting Quadrangular Text}. We evaluate the proposed PSENet on the ICDAR 2015 and ICDAR 2017 MLT datasets to test its ability of detecting the oriented quadrangular text. 
% We train PSENet using ICDAR 2015 and ICDAR 2017 MLT datasets together. 
During testing, we resize the longer side of input images to 2240 and 3200 for ICDAR 2015 and ICDAR 2017 MLT, respectively. For fair comparisons, we report all the single-scale results on these two datasets.

\begin{figure}[t]
  \centering
  \setlength{\fboxrule}{0pt}
  \fbox{\includegraphics[width=1\textwidth]{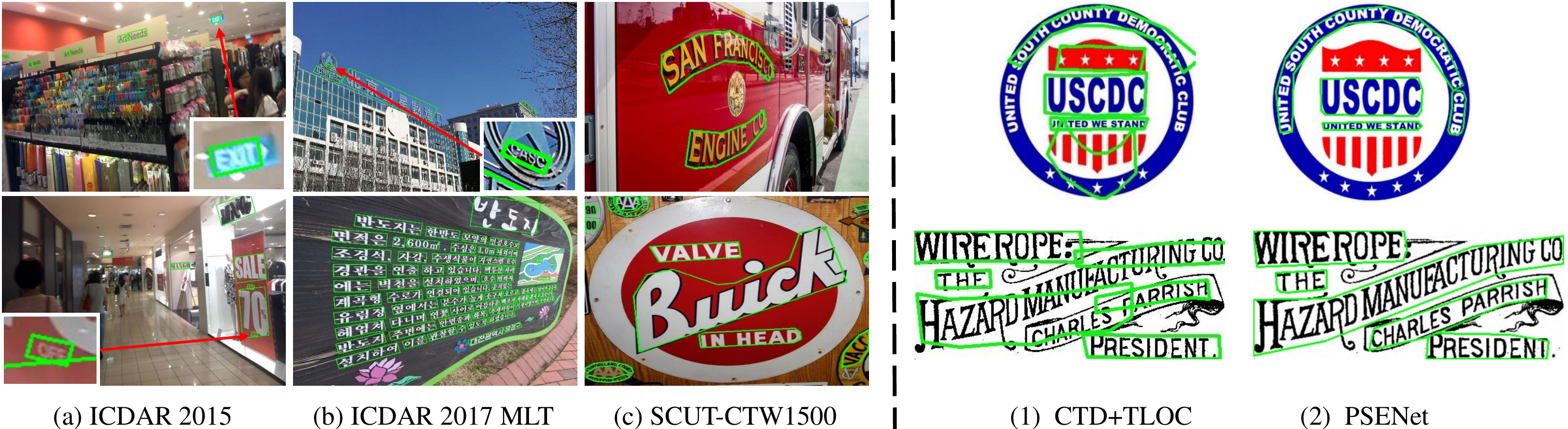}}
  \vspace{-12pt}
  \caption{PSENets' results on three benchmarks and several representative comparisons of curve texts on SCUT-CTW1500. More examples are provided in the \textbf{supplementary materials}.}
  \label{fig:res}
  \vspace{-8pt}
\end{figure}

We compare our method with other state-of-the-art methods in Table \ref{tab:ic-ctw}. Our method outperforms almost all the state-of-the-art methods in the aspect of F-measure. On ICDAR 2015, PSENet achieves the best recall (85.51\%) and obtains the comparable F-measure with the FOTS \cite{liu2018fots}. Furthermore, our results on ICDAR 2017 MLT are even more encouraging with the best F-measure (72.45\%), which surpass the second best method FOTS \cite{liu2018fots} by absolute 5.2\%. The competitive results on the both ICDAR datasets validate the effectiveness of PSENet in the mainstream quadrangular text detection tasks. In addition, we demonstrate some test examples in Fig. \ref{fig:res} (a) (b), and PSENet can accurately locate the text instances with various orientations. Furthermore, we also test the speed (FPS) on NVIDIA GTX 1080 Ti (Table \ref{tab:ic-ctw}) and confirm the satisfactory efficiency of PSENet. 

\begin{table}[h]
	\scriptsize
	% \small
	\centering
	\renewcommand\arraystretch{1.2}
	\newcommand{\tabincell}[2]{\begin{tabular}{@{}#1@{}}#2\end{tabular}}
	\caption{The single-scale results on ICDAR 2015, ICDAR 2017 MLT and SCUT-CTW1500. ``P'', ``R'' and ``F'' refer to precision, recall and F-measure respectively. * indicates the results from \cite{Liu2017Detecting}. ``1s'', ``2s'' and ``4s'' means the width and height of the output map are $1/1$, $1/2$ and $1/4$ of the input test image. The best, second-best F-measure are highlighted in red and blue, respectively.}
	\vspace{-2pt}
	\scalebox{1}{
		\begin{tabular}{c|c|c|c|c|c|c|c|c|c|c}
			\hline
			\multirow{2}{*}{Method} & \multicolumn{4}{c|}{IC15} & \multicolumn{3}{c|}{IC17-MLT} & \multicolumn{3}{c}{SCUT-CTW1500} \\
			\cline{2-11}
			& P (\%) & R (\%) & F (\%) & FPS & P (\%) & R (\%) & F (\%) & P (\%) & R (\%) & F (\%)\\
			\hline
			CTPN \cite{tian2016detecting} & 74.22 & 51.56 & 60.85 & 7.1 & - & - & - & 60.4* & 53.8* & 56.9* \\
			\hline
			SegLink \cite{shi2017detecting} & 74.74 & 76.50 & 75.61 & - & - & - & - & 42.3* & 40.0* & 40.8* \\
			\hline
			SSTD \cite{he2017single} & 80.23 & 73.86 & 76.91 & 7.7 & - & - & - & - & - & - \\
			\hline
			WordSup \cite{hu2017wordsup} & 79.33 & 77.03 & 78.16 & - & - & - & - & - & - & - \\
			\hline
			EAST \cite{zhou2017east} & 83.27 & 78.33 & 80.72 & 6.52 & - & - & - & 78.7* & 49.1* & 60.4* \\
			\hline
			R$^2$CNN \cite{jiang2017r2cnn} & 85.62 & 79.68 & 82.54 & - & - & - & - & - & - & - \\
			\hline
			FTSN \cite{dai2017fused} & 88.65 & 80.07 & 84.14 & - & - & - & - & - & - & - \\
			\hline
			SLPR \cite{zhu2018sliding} & 85.5 & 83.6 & 84.5 & - & - & - & - & 80.1 & 70.1 & 74.8 \\
			\hline
			linkage-ER-Flow \cite{icdar2017mlt} & - & - & - & - & 44.48 & 25.59 & 32.49 & - & - & - \\
			\hline
			TH-DL \cite{icdar2017mlt} & - & - & - & - & 67.75 & 34.78 & 45.97 & - & - & - \\
			\hline
			TDN SJTU2017 \cite{icdar2017mlt} & - & - & - & - & 64.27 & 47.13 & 54.38 & - & - & - \\
			\hline
			SARI FDU RRPN v1 \cite{icdar2017mlt} & - & - & - & - & 71.17 & 55.50 & 62.37 & - & - & - \\
			\hline
			SCUT DLVClab1 \cite{icdar2017mlt} & - & - & - & - & 80.28 & 54.54 & 64.96 & - & - & - \\
			\hline
			Lyu et al. \cite{lyu2018multi} & 89.5 & 79.7 & 84.3 & 3.6 & 83.8 & 55.6 & 66.8 & - & - & - \\
			\hline
			FOTS \cite{liu2018fots} & 91.00 & 85.17 & \textcolor{red}{87.99} & 7.5 & 80.95 & 57.51 & 67.25 & - & - & - \\
			\hline
			CTD+TLOC \cite{Liu2017Detecting} & - & - & - & - & - & - & - & 77.4 & 69.8 & 73.4 \\
			\hline
			\hline
			PSENet-4s (ours) & 87.98 & 83.87 & 85.88 & 12.38 & 75.98 & 67.56 & 71.52 & 80.49 & 78.13 & 79.29 \\
			\hline
			PSENet-2s (ours)& 89.30 & 85.22 & \textcolor{blue}{87.21} & 7.88 & 76.97 & 68.35 & \textcolor{blue}{72.40} & 81.95 & 79.30 & \textcolor{blue}{80.60} \\
			\hline
			PSENet-1s (ours)& 88.71 & 85.51 & 87.08 & 2.33 & 77.01 & 68.40 & \textcolor{red}{72.45} & 82.50 & 79.89 & \textcolor{red}{81.17} \\
			\hline
			
	\end{tabular}}
	\label{tab:ic-ctw}
	% \vspace{-10pt}
\end{table}

\textbf{Detecting Curve Text}. To test the ability of detecting arbitrarily shaped text, we evaluate our method on SCUT-CTW1500, which mainly contains the curve texts. In test stage, we resize the longer side of images to 1280 and evaluate the results using the same evaluation method with \cite{Liu2017Detecting}. We report the single-scale performance on SCUT-CTW1500 in Table \ref{tab:ic-ctw}, in which we can find that the precision (82.50\%), recall (79.89\%) and F-measure (81.17\%) achieved by PSENet significantly outperform the ones of other competitors. Remarkably, it surpasses the second best record by 6.37\% in F-measure. The result on SCUT-CTW1500 demonstrates the solid superiority of PSENet when detecting curve or arbitrarily shaped texts. We also illustrate several challenging results in Fig. \ref{fig:res} (c) and make some visual comparisons to the state-of-the-art CTD+TLOC \cite{Liu2017Detecting} in Fig. \ref{fig:res} (1) (2). The comparisons clearly demonstrate that PSENet can successfully distinguish very complex curve text instances.

\iffalse
\begin{table}[t]
	\scriptsize
	% \small
	\centering
	\renewcommand\arraystretch{1.2}
	\newcommand{\tabincell}[2]{\begin{tabular}{@{}#1@{}}#2\end{tabular}}
	\caption{The experiment result on SCUT-CTW1500. The best, second-best F-measure are highlighted in red and blue, respectively.}
	% \vspace{-10pt}
	\scalebox{1}{
		\begin{tabular}{c|c|c|c}
			\hline
			Method & Precision & Recall & F-measure \\
			\hline
			Seglink [*] & 42.3 & 40.0 & 40.8 \\
			\hline
			CTPN [*] & 60.4 & 53.8 & 56.9 \\
			\hline
			EAST [*] & 78.7 & 49.1 & 60.4 \\
			\hline
			CTD [*] & 74.3 & 65.2 & 69.5 \\
			\hline
			CTD+TLOC [*] & 77.4 & 69.8 & 73.4 \\
			\hline
			SLPR [*] & 80.1 & 70.1 & \textcolor{blue}{74.8} \\
			\hline
			\hline
			PSENet & 82.50 & 79.89 & \textcolor{red}{81.17} \\
			\hline
		\end{tabular}}
	\label{tab:ctw1500}
	% \vspace{-10pt}
\end{table}
\fi

\subsection{More Comparisons on SCUT-CTW1500}
\begin{figure}[h]
	%\vskip -0.2in
	\begin{center}
		{\includegraphics[width=1\columnwidth]{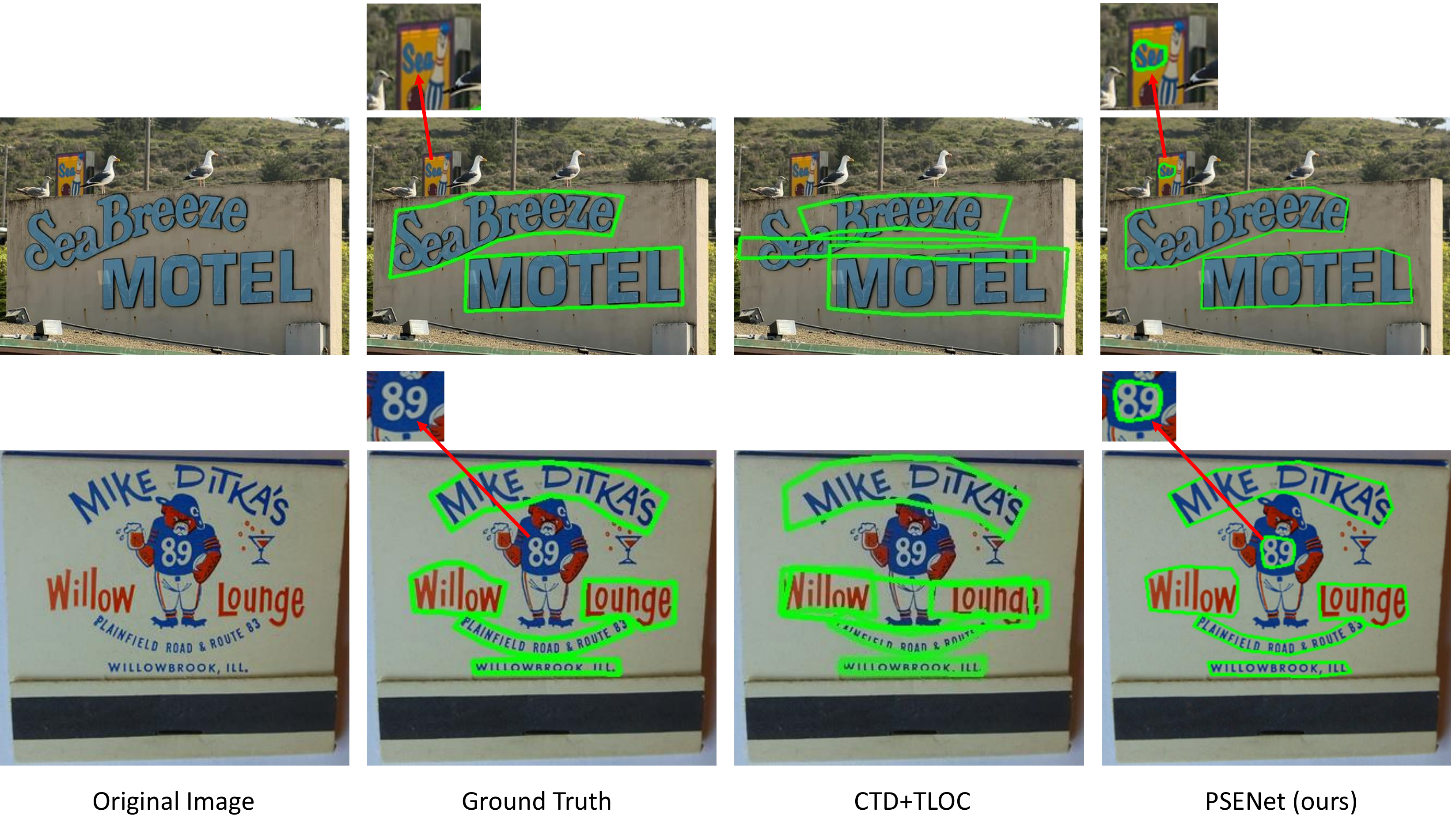}}
		\vspace{-10pt}
		\caption{Comparisons on SCUT-CTW1500. The proposed PSENet produces several detections that are even missed by the groundtruth labels. }
		\label{fig_ctw_nb_cropped}
	\end{center}
\end{figure}

\begin{figure}[t]
	%\vskip -0.2in
	\begin{center}
		{\includegraphics[width=1\columnwidth]{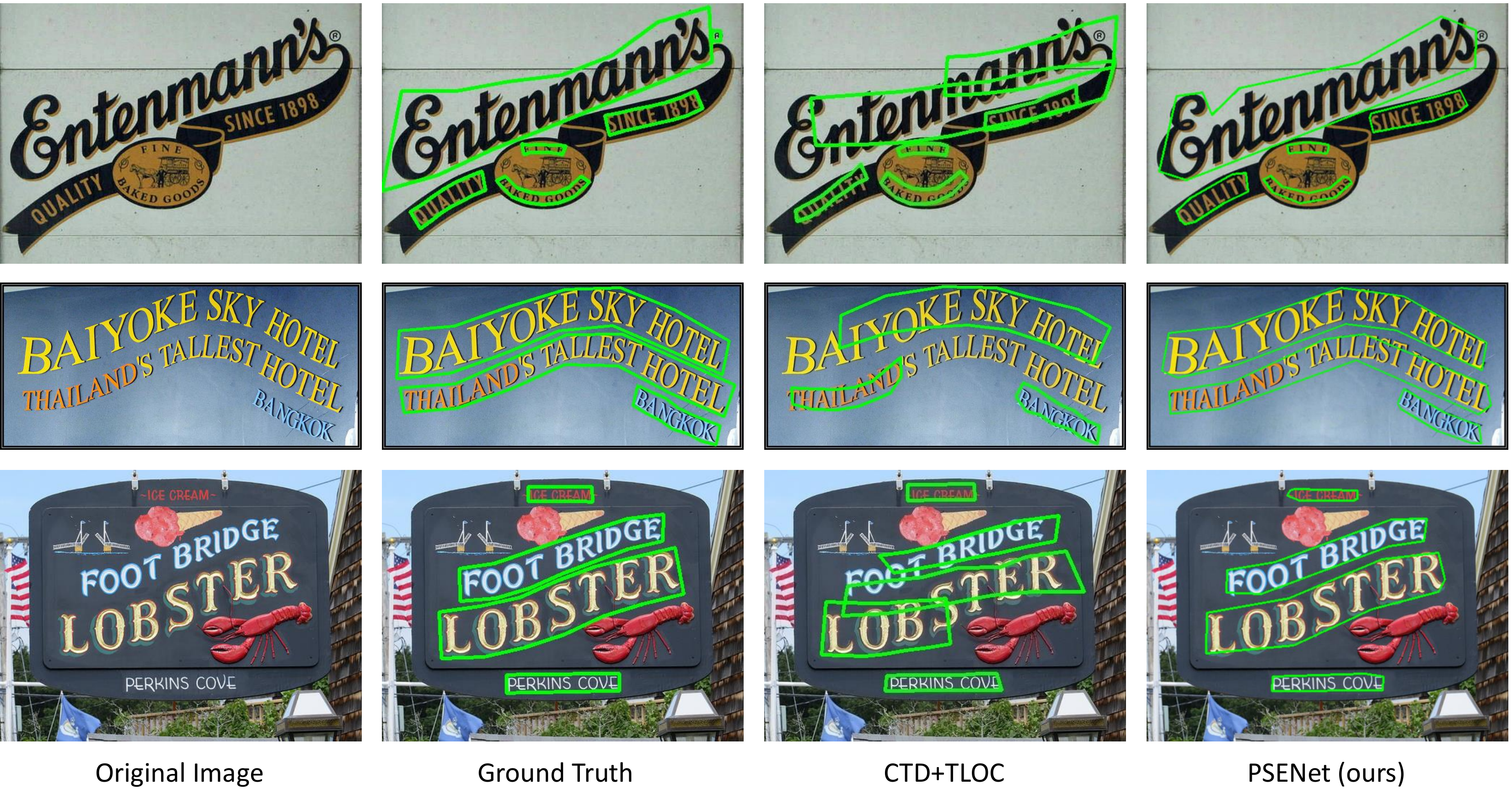}}
		\vspace{-10pt}
		\caption{Comparisons on SCUT-CTW1500.}
		\label{fig_ctw_cropped}
	\end{center}
\end{figure}

\begin{figure}[t]
	%\vskip -0.2in
	\begin{center}
		{\includegraphics[width=1\columnwidth]{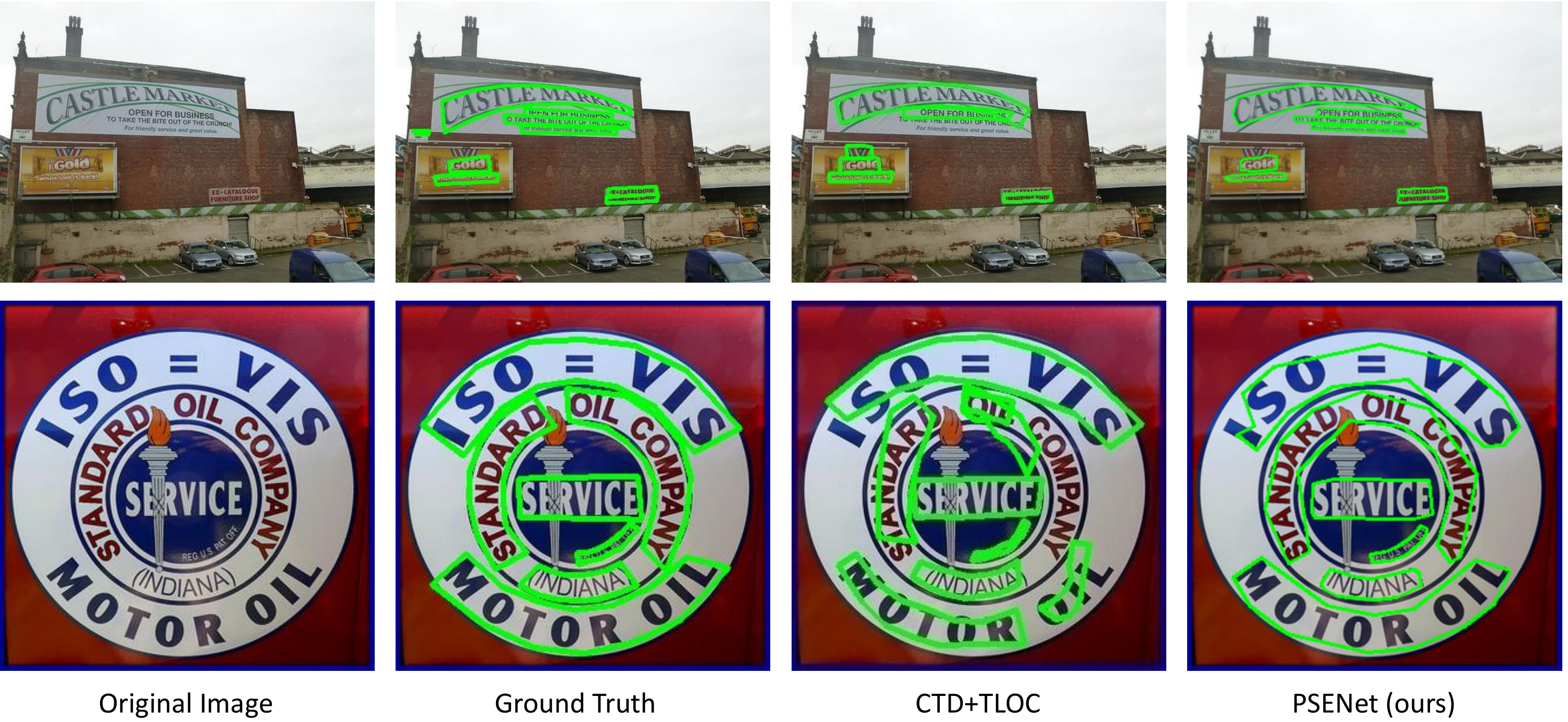}}
		\vspace{-10pt}
		\caption{Comparisons on SCUT-CTW1500.}
		\label{fig_ctw2_cropped}
	\end{center}
\end{figure}

In this section, we show more comparisons on SCUT-CTW1500 in Fig. \ref{fig_ctw_nb_cropped}, \ref{fig_ctw_cropped}, \ref{fig_ctw2_cropped}. It is interesting and amazing to find that in Fig. \ref{fig_ctw_nb_cropped}, our proposed PSENet is able to locate several text instances where the groundtruth labels are even unmarked. This highly proves that our method is quite robust due to its strong learning representation and distinguishing ability. Fig. \ref{fig_ctw_cropped} and \ref{fig_ctw2_cropped} demonstrate more examples where PSENet can not only detect the curve text instances even with extreme curvature, but also separate those close text instances in a good manner.

\subsection{More Detected Examples on ICDAR 2015 and ICDAR 2017 MLT}
\begin{figure}[t]
	%\vskip -0.2in
	\begin{center}
		{\includegraphics[width=1\columnwidth]{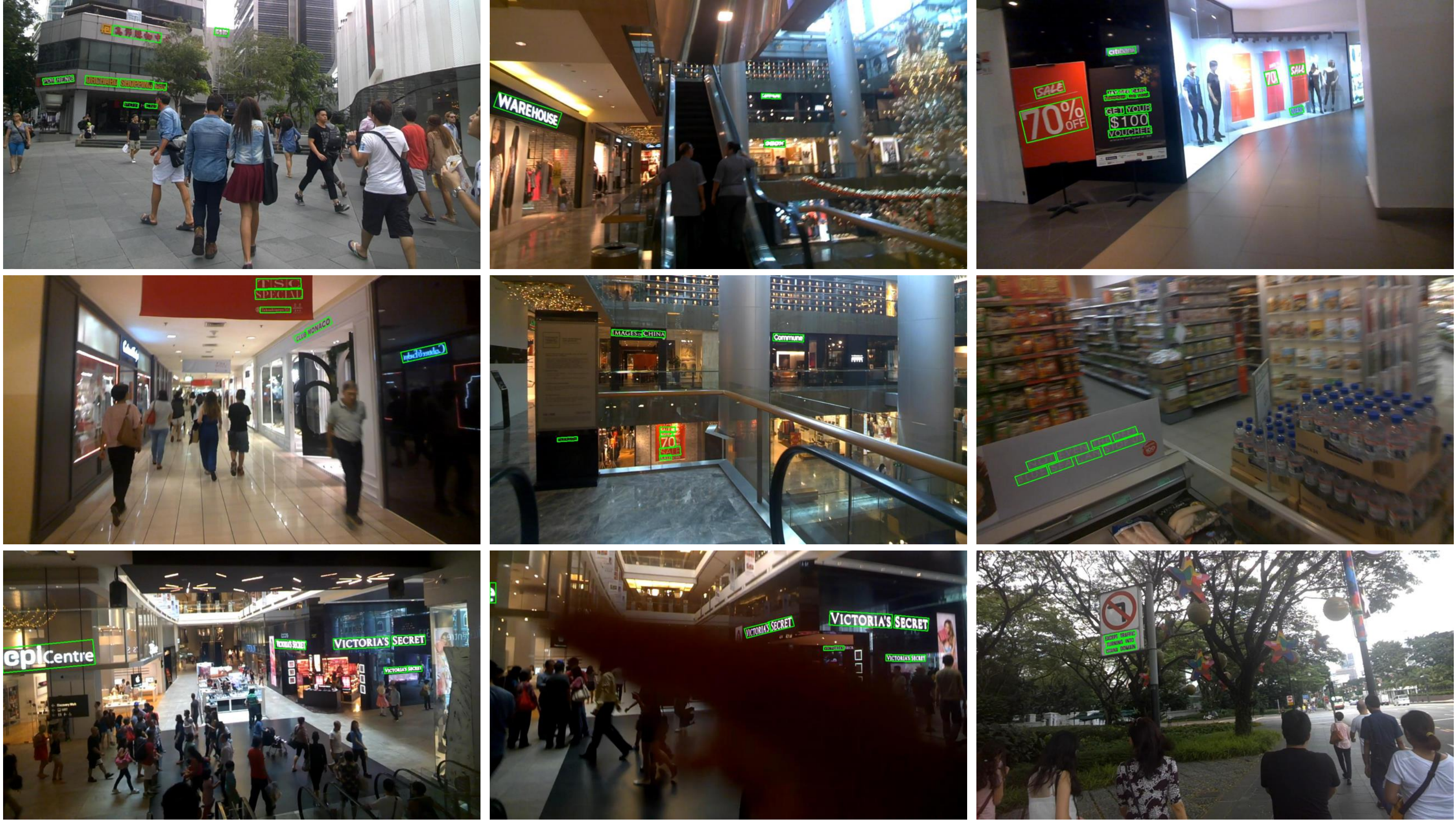}}
		\vspace{-10pt}
		\caption{Test examples on ICDAR 2015 produced by PSENet.}
		\label{fig_ic15_cropped}
	\end{center}
\end{figure}

\begin{figure}[t]
	%\vskip -0.2in
	\begin{center}
		{\includegraphics[width=1\columnwidth]{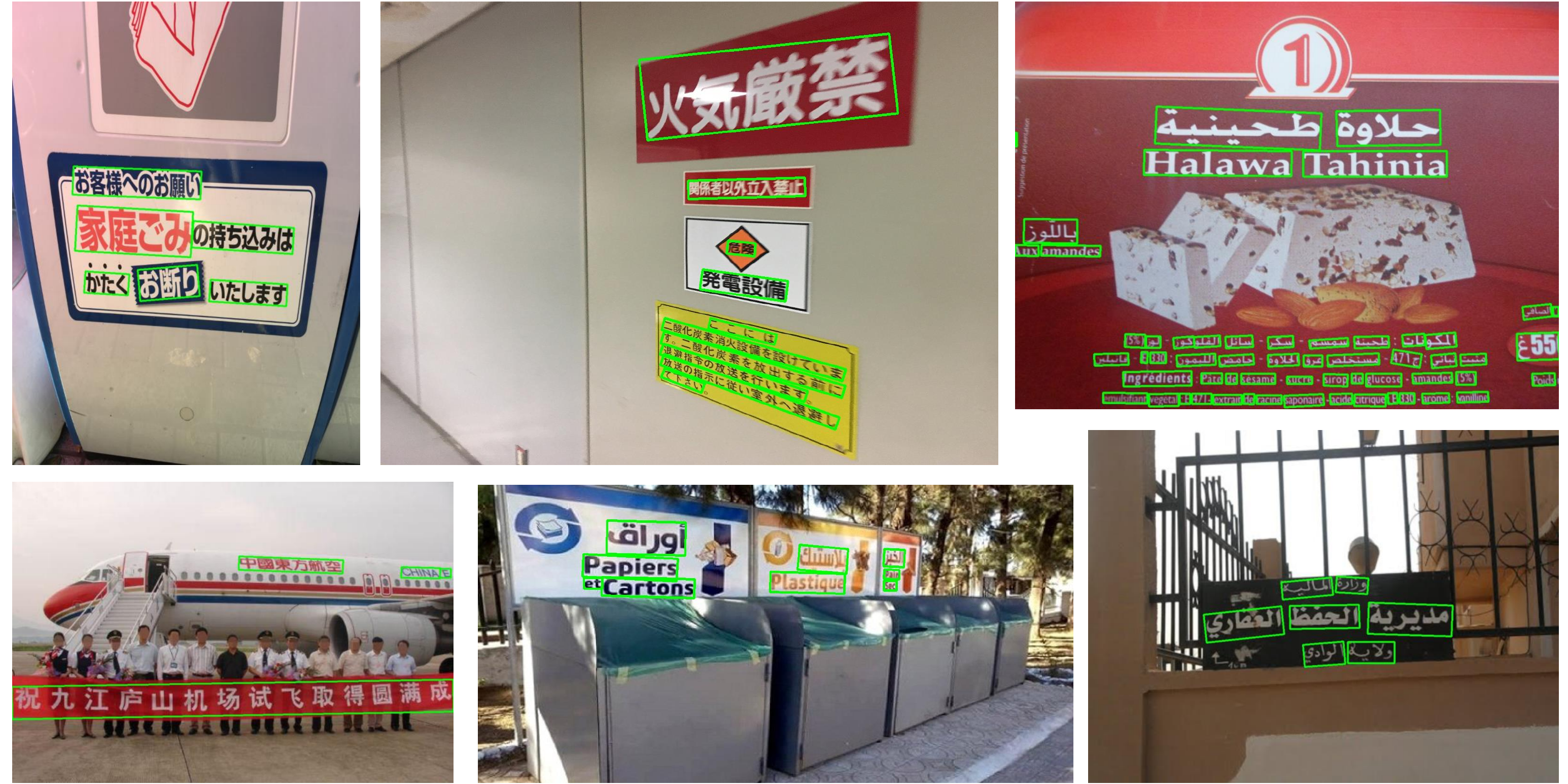}}
		\vspace{-10pt}
		\caption{Test examples on ICDAR 2017 MLT produced by PSENet.}
		\label{fig_ic17_cropped}
	\end{center}
\end{figure}

In this section, we demonstrate more test examples produced by the proposed PSENet in Fig. \ref{fig_ic15_cropped} (ICDAR 2015) and Fig. \ref{fig_ic17_cropped} (ICDAR 2017 MLT). From these results, it can be easily observed that with the progressive scale expansion mechanism, our method is able to separate those text instances that are very close to each other, and it is also very robust to various orientations. Meanwhile, thanks to the strong feature representation, PSENet can as well locate the text instances with complex and unstable illumination, different colors and variable scales.

\section{Conclusion and Future Work}
We propose a novel Progressive Scale Expansion Network (PSENet) to successfully detect the text instances with arbitrary shapes in natural scene images. By gradually expanding the detected areas from small kernels to large and complete instances via multiple semantic segmentation maps, our method is robust to shapes and can easily distinguish those text instances which are very close or even partially intersected. The experiments on scene text detection benchmarks demonstrate the superior performance of the proposed method. 

There are multiple directions to explore in the future. Firstly, we will investigate whether the scale expansion algorithm can be trained along with the network in an end-to-end manner. Secondly, the progressive scale expansion algorithm can be introduced to the general instance-level segmentation tasks, especially in those bencnmarks with many crowded object instances. We are cleaning our codes and will release them soon.

\bibliographystyle{plain}
\bibliography{nips_2018}

\end{document}